\documentclass{elsart1}

\usepackage{graphicx}
\usepackage{amssymb}
\usepackage{amsmath}
\graphicspath{{eps/}}
\usepackage{caption}
\usepackage{subfig}
\usepackage{cite}

\begin{document}
\bibliographystyle{ieeetr}
\begin{frontmatter}

\title{Delay Learning Architectures for Memory and Classification}

\author[label1]{Shaista Hussain\corauthref{cor1}},
\ead{shaista001@e.ntu.edu.sg}
\corauth[cor1]{Corresponding author}
\author[label1]{Arindam Basu},
\author[label2]{R. Wang} and
\author[label2,label3]{Tara Julia Hamilton}
\address[label1]{School of Electrical and Electronic Engineering, Nanyang Technological University, 50 Nanyang Avenue, Singapore 639798}
\address[label2]{University of Western Sydney, Penrith, NSW 2751, Australia}
\address[label3]{School of Electrical Engineering and Telecommunications, University of New South Wales, Sydney, NSW 2052, Australia}

\begin{abstract}
We present a neuromorphic spiking neural network, the DELTRON, that can remember and store patterns by changing the delays of every connection as opposed to modifying the weights. The advantage of this architecture over traditional weight-based ones is simpler hardware implementation without multipliers or digital-analog converters (DACs) as well as being suited to time-based computing. The name is derived due to similarity in the learning rule with an earlier architecture called Tempotron. The DELTRON can remember more patterns than other delay-based networks by modifying a few delays to remember the most `salient' or synchronous part of every spike pattern. We present simulations of memory capacity and classification ability of the DELTRON for different random spatio-temporal spike patterns. The memory capacity for noisy spike patterns and missing spikes are also shown. Finally, we present SPICE simulation results of the core circuits involved in a reconfigurable mixed signal implementation of this architecture.
\end{abstract}


\end{frontmatter}

\section{Introduction: Delay-based Learning Approach}
Neuromorphic systems emulate the behavior of biological nervous systems with the primary aims of providing insight into computations occurring in the brain as well as enabling artificial systems that can operate with human-like intelligence at power efficiencies close to biological systems. Though initial efforts were mostly limited to sensory systems \cite{schaik_cochlea,tobi_imager}, the focus of research has slowly shifted towards the implementation of functions of higher brain regions like recognition, attention, classification etc. However, most of the previous researchers have primarily focused on the implementations of somatic nonlinearity, compact learning synapses and address event representation (AER) for asynchronous communication \cite{Indiveri2011,Bartolozzi2007,biofpaa_jrnl_my,iscas_bif_my,brink_learningfg,nullcline_neu_my,elm_basu_spiking}. As a result, there is a need for modeling and understanding the computational properties of other components of our neurons: the axons and dendrites which have been largely ignored till now. This is also facilitated by recent experimental and computational work which has shed light on possible computational roles of these structures.

The research on spiking neural networks has led to the emergence of a new paradigm in neural networks, which has garnered a lot of interest lately. Several recent studies have presented spiking neural networks to implement biologically consistent neural and synaptic mechanisms \cite{snn_habit,snn_hardware,curr_syn}. As shown by Izhikevich, spiking neural networks with axonal delays have immense information capacity \cite{polychronization}. These networks can exhibit a large number of stereotypical spatio-temporal firing patterns through a combination of spike timing dependent plasticity (STDP) and axonal propagation delays. Learning schemes based on modifying delays can be utilized to read out these firing patterns. This has spurred a renewed interest in the possible role of delays and has even led to analog VLSI implementations of delay models of axons \cite{delay_ckt,mark_axon2}. In this paper we present a computational model-DELTRON which can learn spatio-temporal spike patterns by modifying the delay associated with the spikes arriving at a synaptic afferent. Compared to most earlier implementations \cite{delay_ckt,mark_axon2} that need `N' delay storage elements to memorize a single `N' dimensional pattern, we show much increased memory capacity by modifying only a few delays to memorize the most `salient' part of each pattern. Here `salient' refers to that part of a spatio-temporal pattern which has maximum synchrony or the largest number of coincidental spikes when observed at the soma of the post-synaptic neuron. The synchronous activity of the neurons has been linked to a variety of cognitive functions. Therefore, delay adaptation, which utilizes the synchrony in spike patterns, can play a role in object recognition, attention and neuronal communication.

In the past, several delay learning schemes have been presented for non-spiking networks \cite{delay_dyn,tempo2,delay_pnas} and some of them have been used in applications like word recognition \cite{delay_speech}. In the context of spiking neurons and pulse coupled networks, delay adaptation was implemented in \cite{delay_bpn1,delay_bpn2} for biologically motivated networks using standard analog hardware elements. The delay learning rule for recognizing impulse patterns is similar to our method except that in these studies, the delay parameters are adjusted until all the impulses are coincident while we modify only a subset of delays corresponding to the most synchronous spikes. Our learning rule, initially presented in \cite{shaista_apccas}, is similar to the one presented in \cite{delay_neco} with two differences: we do not have the nonlinear membrane voltage dependent weighting term and we use a single time-based delay adjustment instead of an integral over a time period. More importantly, there is no discussion on the memory capacity of such networks in \cite{delay_neco} with the authors having demonstrated the memorization of a single pattern only.

This paper is organized as follows: introduction to delay-based learning approaches is given after which, the first section presents the computational architecture of DELTRON followed by the learning algorithm in the next section. The fourth section presents simulation results. We discuss details of an efficient mixed-signal VLSI implementation of this algorithm in the fifth section and follow it with conclusions in the last section.
\begin{figure}[!t]
\centerline{\includegraphics[width=0.9\textwidth]{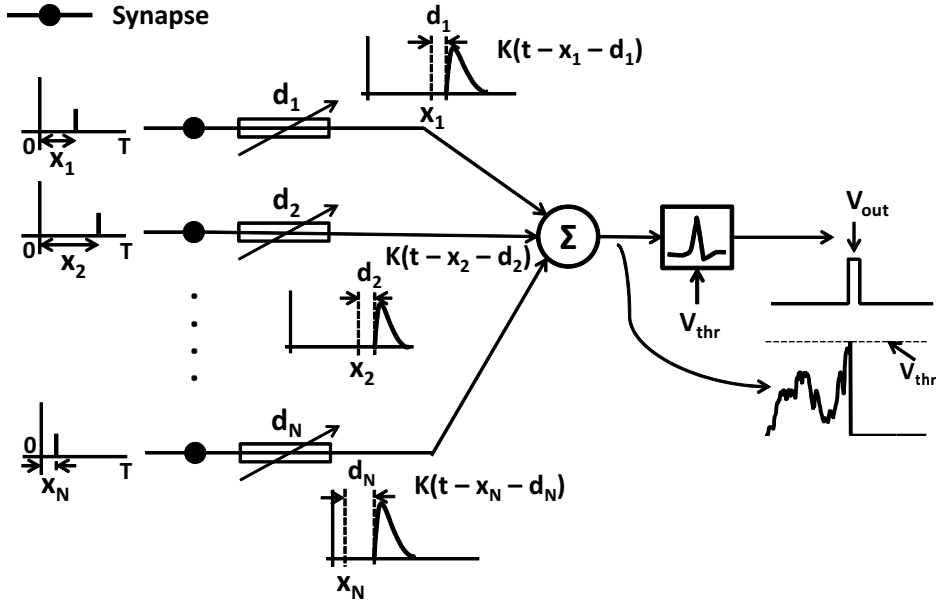}}
\caption{Delay-based learning model where $N$ synaptic afferents receive incoming spikes fired at time $x_i$ and create EPSP waveforms delayed by $d_i$, i = 1, 2, ..., N. Spike delays $\bf d$ = $(d_1, d_2, d_3,\ldots d_N)$ are modified such that membrane potential V(t) crosses the $V_{thr}$.}
\label{fig:architecture}
\end{figure}
\section{The DELTRON Model}
\subsection{Network Architecture}
Figure \ref{fig:architecture} depicts the architecture of the DELTRON that comprises an integrate and fire (I\&F) neuron at the output and $N$ excitatory synapses that receive spiking inputs. Each of these incoming spikes create a delayed excitatory post-synaptic potential (EPSP) that gets linearly summed at the soma. In the bio-physical world, such delays could be attributed to synaptic processes \cite{delay_neco} or dendritic propagation times \cite{hmm_dendrite}. If the summed membrane potential crosses a threshold, $V_{thr}$, the I\&F neuron generates a spike and resets the membrane voltage \cite{GerstnerAug2002}. We want to develop a learning rule that can modify the delays associated with each input so that only a certain desired set of $P$ patterns can fire the neuron by making the membrane potential cross the threshold.

In this paper, we consider applying the DELTRON to classifying or memorizing patterns in the case where there is exactly one spike on each input $i$ at a random time $x_i$ within a fixed time period $T$, i.e, $x_i\in [1$ $T]$, i = 1, 2, ..., N. This case corresponds to applying the DELTRON to classify signals coming from a sensor employing time-to-first-spike (TTFS) encoding \cite{ttfs_imgSen,ttfs_Shoushun,ttfs_imager}. Time based encoding is becoming popular recently due to the reduced supply voltage (leading to lower voltage headroom) and increased speeds (leading to higher temporal resolution) in today's deeply scaled VLSI processes; hence, the DELTRON will be very useful as a back end processor for all such temporal encoding sensory systems. Formally, we can express the membrane voltage $V(t)$ as a sum of the EPSPs generated by all incoming spikes as:
\begin{align}
V(t)=\sum_{i}K(t-t_{i})
\end{align}
where $K:R\rightarrow R$ is the EPSP kernel function, $d_i$ is the delay of the i-th branch and $t_i=x_i+d_i$, i = 1, 2, ..., N. The vector $\bf x$ = ($x_1$, $x_2$, ..., $x_N$) constitute a spike pattern presented to the network. In this work, we consider the fast rising and slowly decaying PSP kernel to be given by $K(t) = V_0(exp[-(t)/\tau] - exp[-(t)/\tau_s])$, where $\tau$ is the synaptic current fall time constant and $\tau_s$ the synaptic current rise time constant. Our analysis is quite general and is applicable to other forms of the function $K$ as well. As mentioned earlier, the I\&F output neuron elicits a spike when the voltage $V(t)$ crosses the threshold voltage $V_{thr}$. Let $n_{spk}$ denote the number of spikes fired by the output neuron for the presentation of a pattern. Then, the operation of the neuron is described as:
\begin{align}
&\textrm{If } V(t)>V_{thr},\notag\\
&V(t)\rightarrow 0\notag\\
&n_{spk}(t)=n_{spk}(t-1)+1 (n_{spk}(0)=0)
\end{align}
The final output of the network, $y$, is a logical variable having a value of $1$ if the pattern is recognized. We define this operation by:
\begin{align}
y&=1 \textrm{ if }n_{spk,final}>0 \notag\\
 &=0 \textrm{ otherwise}
\end{align}	
where $n_{spk,final}$ is the final value of $n_{spk}$ after the presentation of the pattern is completed. In other words, we declare the pattern recognized if the neuron fires one or more spikes in the entire duration.
\subsection{Input Pattern Space}
 We mentioned earlier that the input spike pattern to the network is $\bf x$ = ($x_1$, $x_2$, ..., $x_N$) where $x_i\in [1$ $T]$, i = 1, 2, ..., N. For any real world inputs, there is a finite precision $\Delta t$ at which an input $x_i$ can change. Without loss of generality, let us assume that $\Delta t=1$ so that $x_i$ are integers (otherwise, we can always re-define $T'=T\diagup\Delta t$ and have $x_i \in[1$ $T']$ be integers). Then, the total number of possible patterns, $P_T$ is given by $P_T=T^N$. However, some of these patterns are very similar to each other and should be considered to belong to the same category. For example, a TTFS imager presented with the same scene will produce slightly different spike times on different trials due to noise inherent in such systems. Hence, we define two patterns $x$ and $y$ to belong to the same category if:
\begin{align}
||x-y||_\infty\leq s,s\in N
\end{align}
This results in a total of $(2s+1)^N$ patterns in each category. In some applications, sometimes spikes may be missing on several inputs. To accommodate such cases, we can define a pattern to also belong to a category if it is exactly same as one of the existing patterns in that category but has $m$ missing spikes. Then, the total number of patterns in each category for cases where the maximum number of missing spikes is $M$ is given by $P_{cat}=\sum_{m=0}^{M}{^N}C_m(2s+1)^N$, where ${^N}C_m$ is the total number of ways in which $m$ out of the total $N$ spikes may be missing. This leads to the total number of categories being $P_C=P_T\diagup P_{cat}$. This provides an upper bound on the total number of patterns $P$ that the network can be trained to recognize. For the parameter values we use ($s<<T$, $N>>1$, $P<<P_C$), the probability of two out of $P$ randomly selected patterns to fall in the same category is extremely small. Hence, unless explicitly mentioned, whenever we mention the capacity of the network to memorize or classify patterns in the rest of the paper, we mean patterns belonging to different categories.

\subsection{Choice of Threshold and the Two types of error}
The DELTRON suffers from two types of errors when used in pattern memorization or classification tasks: false positive (FP) and false negative (FN). Suppose the network is trained to respond to patterns of class 1 by producing $y=1$. In the classification case, it is additionally trained to respond to patterns of class 2 with $y=0$. For pattern memorization, this step is not there and the entire population of patterns excluding those in class 1 act as a class 2 of distractor patterns--hence this is a more difficult task. Now, if a pattern of class 1 is presented during testing and the network responds with $y=0$, it makes a FN error. On the other hand, if a pattern of class 2 is presented and the network produces $y=1$, it is a FP error.

The threshold for the output neuron is an important parameter in deciding the balance between FP and FN errors. This is achieved by first computing the probability distribution of the maximum value ($V_{max}$) of membrane voltage  due to random input spike patterns. As an example, Figure \ref{fig:Vmax}(a) shows $V_{max}$ for a random spike pattern while Figure \ref{fig:Vmax}(b) depicts the probability distribution of $V_{max}$ (parameters of this simulation are given in Section \ref{sec:simulation}). As shown in Figure \ref{fig:Vmax}(b), $V_{thr}$ is set as a value $\Delta V$ units higher than the value $V_{peak}$ at which the probability distribution is maximum, i.e. $V_{thr}=V_{peak}+\Delta V$. It would seem that larger values of $\Delta V$ are better since it reduces FP errors. For example, setting $V_{thr}$ to 16 would almost eliminate FP errors. However, this leads to the problem of `trainability'--the network can no longer successfully memorize a large number of patterns and makes many FN errors. In that case, the optimal threshold ($V_{opt}$) is that which minimizes sum of FP and FN errors as shown in Figure \ref{fig:Vmax}(c). We obtain $V_{max}$ distributions after training the network (red) and when new unseen patterns are presented to the trained network (blue). The $V_{max}$ value at the point of intersection of these two distributions gives us the optimal threshold, $V_{opt}$. The strategy we have taken throughout our work is to keep a large threshold value during training ($V_{thr,TR}=V_{thr}$) and use the optimal threshold during recall.

For the classification case, there is another threshold $V_{thr-}=V_{peak}-\Delta V$ and the network is trained to reduce $V_{max}$ lesser than $V_{thr-}$ for patterns in class 2. Again, the optimal threshold is used during testing. For the sake of simplicity, in the rest of the paper we refer to the training threshold as $V_{thr}$ with the implicit understanding that the optimal threshold is always used during testing.

\section{Learning Algorithms}
\subsection{Delay Learning Algorithm for Pattern Memorization}
\label{sec:algo}
The delay-based learning is similar to the learning in the tempotron \cite{tempotron} except that weights were modified in that case whereas we modify delays. The training set for the memorization task consisted of $P$ input spike patterns and the memory capacity of the network was computed as the number of patterns learnt. A spike pattern is said to be learnt by the network if output $y=1$ in response to that pattern. The model was not trained on distractor patterns for this task. Since the training and testing patterns sets are the same in a memorization task, the problem of overfitting arises. However, in our case the use of low-pass kernel function $K$ provides some degree of resilience to overfitting. The network learns spike patterns by modifying the $N$-dimensional delay vector $\bf d$. The learning algorithm consists of the following steps:

\begin{figure}[!t]
\centerline{\includegraphics[width=1\textwidth]{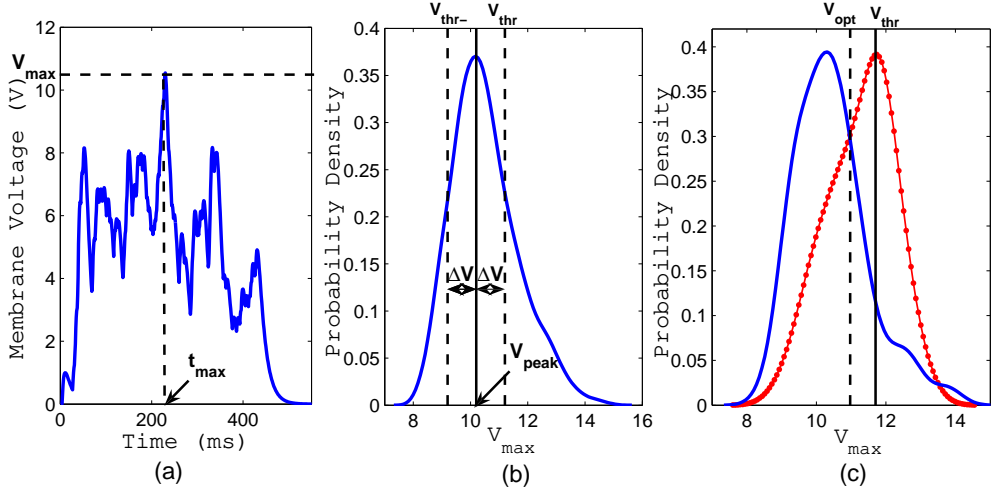}}
\caption{(a) Membrane potential $V(t)$ for a spike pattern. The horizontal line denotes the maximum value of $V(t)$, i.e. $V_{max}$ at time $t_{max}$. (b) Probability density distribution of $V_{max}$ for random spike patterns. $V_{thr}$ is set as $\Delta V$ units greater than $V_{peak}=10.2$ for memorization task and the two thresholds $V_{thr}$ and $V_{thr-}$ are set for classification task. (c) Determination of optimal threshold $V_{opt}$. $V_{max}$ distributions for learnt (red) and new (blue) patterns are obtained. The training threshold $V_{thr}$ used here is $11.7$ corresponding to $\Delta V=1.5$. The point of intersection of the two distribution curves gives $V_{opt}=10.9$ (dashed vertical line).}
\label{fig:Vmax}
\end{figure}

\begin{figure}[!t]
\centerline{\includegraphics[width=0.5\textwidth]{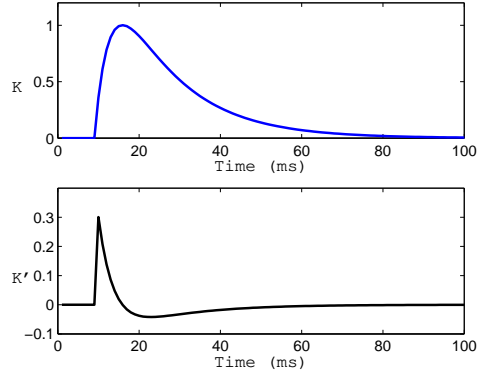}}
\caption{Postsynaptic potential due to a spike arriving at a synapse given by kernel K (top). Derivative of K (bottom).}
\label{fig:Kernel}
\end{figure}

\begin{itemize}
\item[(1)] Delay vector $\bf d$ = $(d_1, d_2, d_3,\ldots d_N)$ is initialized by choosing the value of each $d_i$ from a uniform distribution between $0$ and $d_{init}$.
\item[(2)] In every iteration of the training process, a spike pattern is presented to the network.
\item[(3)] $V(t)$ is computed for the pattern. Maximum value of $V(t)$; $V_{max}$ is found.
\item[(4)] If $V_{max} > V_{thr}$, the spike pattern is learnt. Delays are not modified.
\item[(5)] If $V_{max} < V_{thr}$, time $t_{max}$ at which $V(t)$ attains its maximum value ($V_{max}$) is found.
\item[(6)] The error function is chosen as $E=V_{thr}-V(t_{max})$ since a pattern is learnt if $V(t)$ crosses $V_{thr}$ at least once and $t_{max}$ is the time when it is easiest to do it. $\Delta d$ is calculated using gradient descent on the error function as shown below.
\begin{align}
E &= V_{thr} - V_{max}\\
&= V_{thr} - V(t_{max}) \notag\\
\Delta d_i &= -\frac{\partial E}{\partial d_i} = \frac{\partial V(t_{max})}{\partial d_i} \notag\\
&= \frac{\partial [\sum_j{K(t_{max}-x_j-d_j)}]}{\partial d_i} + \frac{\partial V(t_{max})}{\partial t_{max}} \frac{\partial t_{max}}{\partial d_i} \notag\\
&= - K'(t_{max}-x_i-d_i) \textrm{, since } \frac{\partial V(t_{max})}{\partial t_{max}}=0
\end{align}
where $K'$ indicates the derivative of $K$. Figure \ref{fig:Kernel} shows the PSP kernel K and its derivative.
\item[(7)] Delays are modified according to $\bf d$ = $\bf d$ + $\eta \Delta$$\bf d$ only if the number of learnt patterns increases, where $\eta$ is the learning rate. The boundary values of delay are $0$ and $T$.
\item[(8)] If the number of patterns learnt by the model doesn't increase in 20 consecutive learning iterations, then it is assumed that the algorithm has encountered a local minimum. In an attempt to escape the local minimum, delays are modified in spite of no change in the number of patterns learnt by the network. The last minimum which gave the highest number of learnt patterns is remembered.
\item[(9)] Learning consisting of steps 2-8 is stopped if any of the following conditions occurs
\begin{itemize}
\item[(a)] All patterns are learnt completing the learning process, which is the more frequent outcome when trained on small number of patterns.
\item[(b)] 100 local minima are encountered, which refer to 100 continuous instances of learning algorithm getting trapped in a status of no change in the learning, after which there is no further increase in the number of learnt patterns. This is the more frequent outcome when large number of patterns are being learned. The delays corresponding to the best minimum are the final learnt delays.
\end{itemize}
\end{itemize}
The DELTRON maximizes its memory capacity by modifying only \emph{some} of the delays preferentially over the others. This choice is guided by the spikes arriving within a time window before $t_{max}$. The length of this window is governed by the kernel $K(t)$.

\begin{figure}[!t]
\centerline{\includegraphics[width=0.8\textwidth]{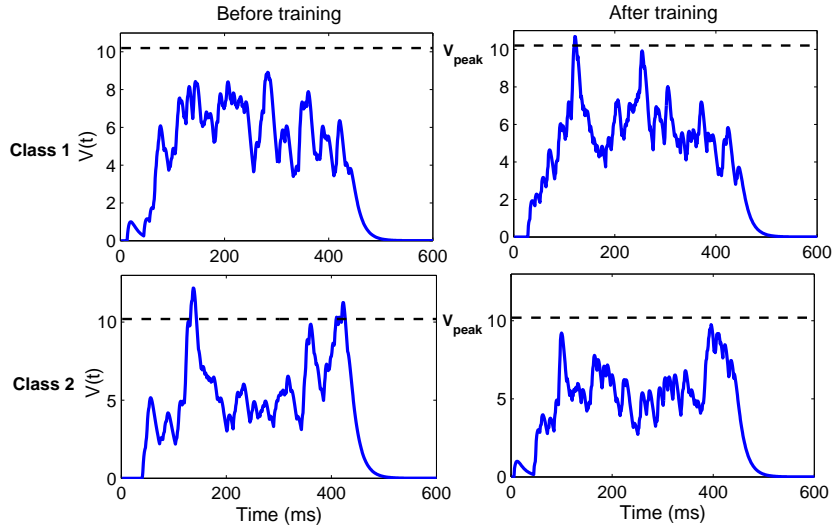}}
\caption{Membrane potential generated by a pattern in class 1 (top row) and in class 2 (bottom row) before and after training show how learning increases $V_{max}$ for the class 1 while reduces it for the class 2. The increased synchrony for class 1 and decreased synchrony for class 2 are shown.}
\label{fig:Class_V}
\end{figure}

\subsection{Delay Learning Algorithm for Pattern Classification}
We have also utilized the delay-modification learning scheme for training our model on a classification task. We generated two sets of random spike patterns, each set consisting of $P$ patterns. Each spike pattern $\bf x$ = ($x_1$, $x_2$, ..., $x_N$) consists of spike times $x_i$ randomly drawn from a uniform distribution between $1$ and $T$ as in the first experiment. These two training sets of spike patterns were arbitrarily assigned to class 1 and class 2. The network was trained such that the output neuron fires at least one spike (network output $y=1$) in response to the patterns of class 1 and fails to fire for patterns in class 2. The learning algorithm is similar to that used in the memory capacity experiment with an extra step for patterns in class 2. The input patterns belonging to the class 1 and class 2 are presented to the network. The delays are modified when a pattern of class 1 is presented and $V_{max} < V_{peak} + \Delta V$ or when a pattern of class 2 is presented and $V_{max} > V_{peak} - \Delta V$. The thresholds for training the patterns in the two classes: $V_{thr}$ for class 1 and $V_{thr-}$ for class 2 are shown in Figure \ref{fig:Vmax}(b). Different $\Delta V$ values were used for training. $\Delta \bf d$ is calculated as before and delays are modified as: $\bf d$ = $\bf d$ + $\eta \Delta$$\bf d$ if the pattern belongs to class 1 and $\bf d$ = $\bf d$ - $\eta \Delta$$\bf d$ if the pattern belongs to class 2. After training is completed, the classification performance of the network is tested by presenting all the patterns on which it was trained. The optimal threshold $V_{opt}$ to test the classification performance of the network is set to $V_{peak}$. Therefore, a pattern in class 1 is correctly classified if it fires the output neuron by satisfying the condition $V_{max}>V_{peak}$ and a pattern in class 2 is correctly classified if it satisfies $V_{max}<V_{peak}$.

This learning process enables us to achieve increased synchrony for patterns in class 1 and decreased synchrony for patterns in class 2. An example of this is shown in Figure \ref{fig:Class_V}. The details of the simulation  setup and parameters are provided in the next section; this figure gives an intuitive feeling for the effect of delay changes on $V(t)$. The delays are modified such that a learnt pattern in class 1 (top), has more synchronized coincident spikes at the soma to allow membrane voltage, $V$ to exceed $V_{peak}$ while a trained pattern in class 2 (bottom) is characterized by asynchronous spikes at the soma, which result in $V$ falling below $V_{peak}$.

\section{Simulation Results}
\label{sec:simulation}
In all the experiments described here, the number of inputs ($N$) is 100, the temporal length of a pattern ($T$) is 400 ms and delays $d_i$ were initialized from a uniform distribution between 0 and 50 ms ($d_{init}=50$ ms). The parameters used for training are as follows: $V_0=2.12$, $\tau=15$ ms, $\tau_s=\tau/4$, initial learning rate $\eta_0=5$ for the first 500 iterations and then reduced by 0.5 after every 500 iterations. $\eta_0$ is chosen such that the maximum change in delay in one iteration is much smaller than $T$. The peak location of the $V_{max}$ distribution was estimated to be $V_{peak}=10.2$, as shown in Figure \ref{fig:Vmax}(b). Different $V_{thr}$ values were used for training by varying $\Delta V$.

\begin{itemize}
\item[(I)] \textbf{Effect of Learning: $V_{max}$ and $t_{max}$ distribution}\\
The first simulation results are shown in Figure \ref{fig:ProbDist}. We obtained the distribution of $V_{max}$ before and after training to see how delay learning changes $V_{max}$ values of spike patterns being learned. The solid blue curves denote the probability distribution of $V_{max}$ for $P=20, 50$ and $100$ spike patterns before the network is trained while the red curves with dots show the distribution after training. There is a clear shift in the peak of the distribution towards $V_{thr}$ indicating that the DELTRON is able to learn to respond preferentially to these spike patterns by shifting the $V_{max}$ values closer to the $V_{thr}$. The black dashed curves depict the response of the trained network to a new random set of $P$ patterns showing that its response to new, not learnt patterns is relatively unchanged. Each row corresponds to the $V_{max}$ distribution for a different $V_{thr}$ value. The $V_{thr}$ values used were relative to the peak value of $V_{max}$ distribution before training as shown in Figure \ref{fig:Vmax}(b). $V_{thr}$ was set to $10.7, 11.2$ and $11.7$ corresponding to $\Delta V = 0.5, 1.0$ and $1.5$ units respectively. We can see from the Figure \ref{fig:ProbDist} that as $V_{thr}$ increases, the separation between the $V_{max}$ distributions before and after training also increases indicating that the network learns the patterns by setting the $V_{max}$ values closer to the $V_{thr}$. For larger values of $V_{thr}$, less number of ``background'' patterns are wrongly recognized to be in the memory and therefore, lower FP errors are incurred.

Figure \ref{fig:tmaxDist} shows the distribution of $t_{max}$, the time at which $V=V_{max}$. As we can see, the $t_{max}$ distribution before delay learning (blue bars) is evenly spread out across time. After training, the $t_{max}$ distribution (red bars) doesn't change considerably which implies that the delays are modified such that the EPSPs due to a spike pattern learn to arrive at times closer to the initial $t_{max}$ value for that pattern, thereby not changing the $t_{max}$ values significantly.

\begin{figure}[!t]
\centerline{\includegraphics[width=1.0\textwidth]{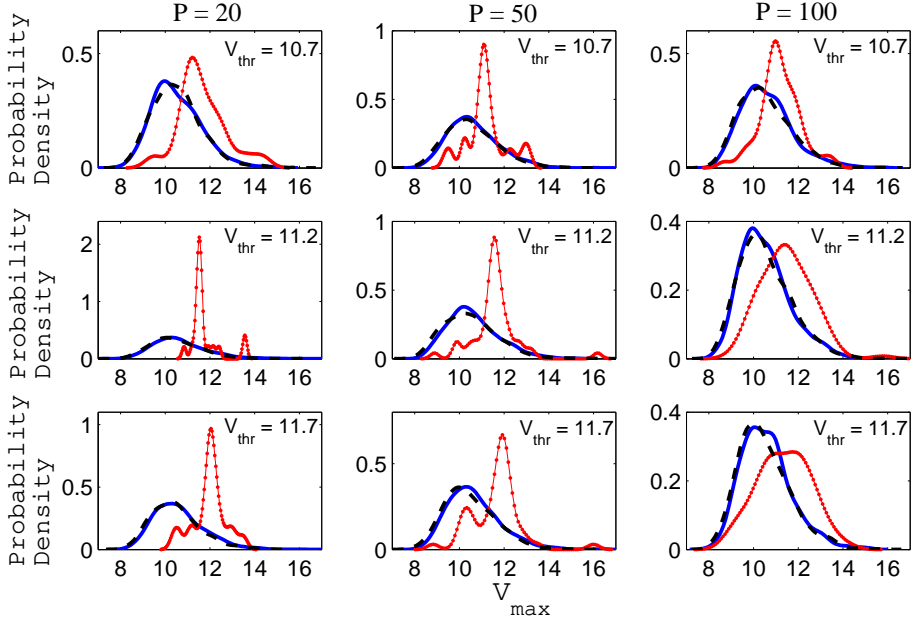}}
\caption{Probability density distribution of $V_{max}$ for a set of 20, 50 and 100 random spike patterns before (blue) and after (red) training showing a clear shift in the peak of the distribution for trained patterns. $V_{max}$ distribution for new patterns presented to the trained network is shown in black.}
\label{fig:ProbDist}
\end{figure}

\begin{figure}[!t]
\centerline{\includegraphics[width=0.6\textwidth]{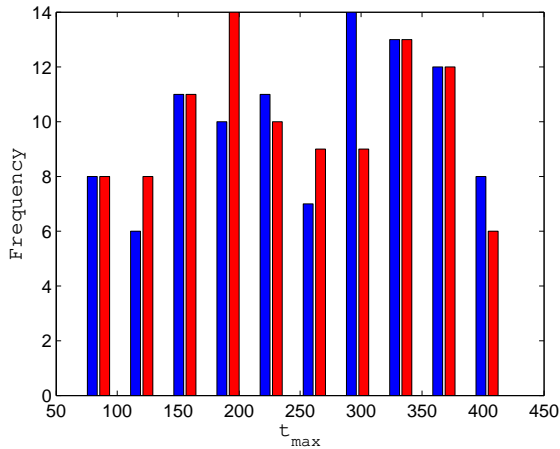}}
\caption{$t_{max}$ distribution for 100 random spike patterns. The $t_{max}$ values before (blue) and after (red) training are spread across the entire pattern duration. $V_{thr}$ = 10.7.}
\label{fig:tmaxDist}
\end{figure}

\item[(II)] \textbf{Pattern Memorization}\\
In the next experiment we calculated the memory capacity of the network. We trained it on $P=10, 20, 30, 50, 70$ and $100$ random spike patterns using different values of $V_{thr}$ as used in the first experiment. $V_{opt}$ values corresponding to a certain value of $P$ and $V_{thr}$ were determined as discussed earlier (Figure \ref{fig:Vmax}(c)) and were used to test how many of the trained patterns are recalled by the network. As shown in the Figure \ref{fig:ProbDist}, the $V_{max}$ distribution for learnt patterns shifts towards $V_{thr}$ used for training. Therefore, as $V_{thr}$ increases, $V_{opt}$, which lies at the point of intersection of the learnt and background $V_{max}$ distributions, also increases. Figure \ref{fig:MemCap} shows the percentage of patterns memorized by the network averaged over 10 repetitions. As discussed earlier, when trained on small number of patterns, $P=20$ and $V_{thr}=10.7$, the learning process completes by memorizing all the patterns, which is the exit condition (a) in the step (9) of learning algorithm. When trained on $P=50$ patterns, the learning algorithm stops as in condition (a) 4 out of 10 times it was trained and terminates by encountering 100 local minima (exit condition (b)) 6 out of 10 times. For large number of patterns, $P=100$, the learning algorithm exits by encountering condition (b) for all the 10 runs. The network memorizes 90-100\% of the patterns when it is trained on small number ($P=10-50$) of patterns with $V_{thr}$ set to 10.7. The maximum number of patterns learned by the network depend on the number of synapses on which spike patterns arrive. In our simulations, the network having $N$=100 afferents can memorize a maximum of about 84\% of the patterns for $P=100$, i.e. about 84 random spike patterns. When higher threshold is used for training, more number of coincident EPSPs are needed for $V_{max}$ to exceed the higher $V_{thr}$, thereby making the delay learning more difficult. This results in less number of learnt patterns and therefore a drop in the memory capacity for higher $V_{thr}$ values, maximum number of patterns memorized by the network reduces from a mean value of 84 to 64 for the highest $V_{thr}=11.7$ used.

Figure \ref{fig:Tot_Err} shows the total error given by the sum of FP and FN errors, incurred by the learning algorithm in the memorization task. We can see that as the number of patterns increases, the total error increases which is mainly due to the higher FN error for larger number of patterns. This has already been shown in Figure \ref{fig:MemCap}, where the memory capacity drops and therefore the FN error increases with the increase in $P$. For small number of patterns, $P=10-20$, the total error is less for higher $V_{thr}$ and therefore higher $V_{opt}$ values. This is due to low FP errors incurred when higher threshold is used to recall the patterns. For moderate and large number of patterns, $P>30$ the total errors are similar across different $V_{thr}$ because as FP error reduces with the increase in $\Delta V$, the FN errors start to rise due to the decreased ability of the network to learn large number of patterns at higher thresholds.

\begin{figure}[!t]
\centerline{\includegraphics[width=0.6\textwidth]{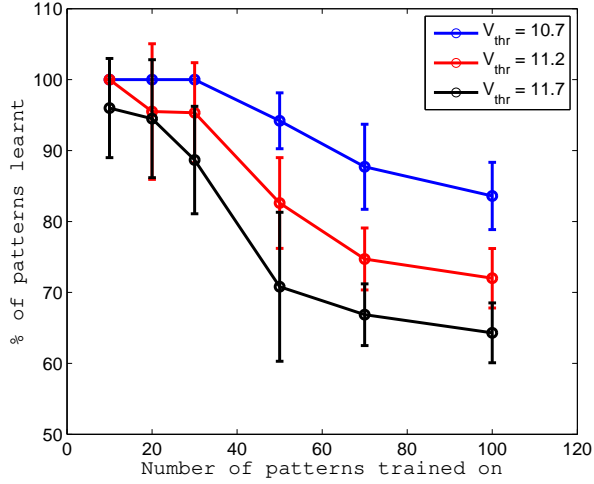}}
\caption{Memory capacity of the model. The percentage of patterns recalled as a function of different number of patterns used to train the network. As the number of patterns used for training increases the fraction of patterns that the network learns reduces. The effect of learning with different $V_{thr}$ values is also shown.}
\label{fig:MemCap}
\end{figure}

\begin{figure}[!t]
\centerline{\includegraphics[width=0.6\textwidth]{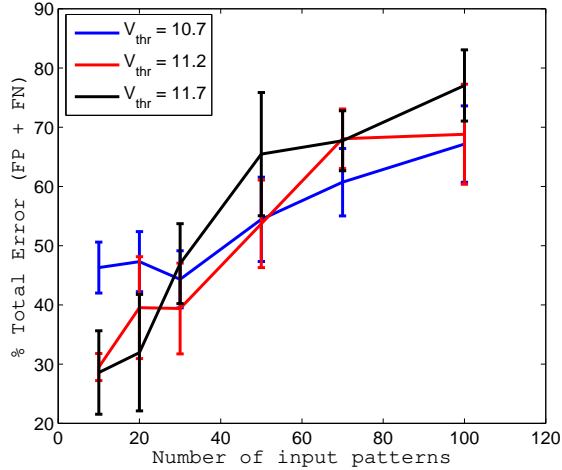}}
\caption{Total error given by the sum of false positive (FP) and false negative (FN) errors in the pattern memorization task as a function of $P$ and the training threshold $V_{thr}$.}
\label{fig:Tot_Err}
\end{figure}

\item[(III)] \textbf{Pattern Classification}\\
The network performance in the classification task was determined by training the network on random spike patterns belonging to class 1 and class 2. As mentioned earlier, a pattern belonging to class 1 is said to be correctly classified if $V_{max}>V_{peak}$ and a class 2 pattern is correctly classified if $V_{max}<V_{peak}$. $V_{peak}=10.2$ and $\Delta V$ values used were 0, 0.2 and 0.4.  The $V_{max}$ distributions for the learnt patterns ($P=50$) in class 1 (red) and class 2 (blue) for different $\Delta V$ are shown in Figure \ref{fig:Class_Dist}. We can see that the separation between the two learnt distributions increases with $\Delta V$, shifting towards the training thresholds $V_{thr}$ and $V_{thr-}$ for the respective classes. When new patterns are presented to the network trained on classification task, the $V_{max}$ distribution (black) obtained is same as the distribution for the learnt patterns before training. As shown in Figure \ref{fig:ClassCap}, 90-100\% patterns are correctly classified when the network is trained on approximately 140 patterns, 70 patterns belonging each to class 1 and class 2. As the number of patterns are increased, the network correctly classifies a maximum of about 80\% of the 200 patterns it is trained on. When $\Delta V$ is increased, the training thresholds for classifying random spike patterns belonging to the two classes shift further away from the peak of the initial $V_{max}$ distribution, increasing for class 1 and decreasing for class 2. In this case, it is expected that the criteria for classifying the patterns in the two classes would be met more easily and therefore, the classification accuracy should be same or higher than the case when $\Delta V=0$. However, the accuracy decreases slightly because higher $V_{thr}$ also imposes a limit on the `trainability' of the network.

As discussed above, the classification ability of the model depends on the number of input patterns $P$ and $\Delta V$. However, this capacity changes with the number of synapses $N$. Therefore, we investigated the capacity of DELTRON as a function of the parameter $\alpha=P/N$, where $\alpha$ is a measure of the load on the system \cite{tempotron}. We tested the performance of the network in classifying spike patterns for different values of $N=50, 100$ and $200$ by varying the number of patterns for each case and $\Delta V=0$. As shown in Figure \ref{fig:Capacity_Load}, the classification capacity is plotted with respect to $\alpha$ for different $N$. We can see that the model can learn to classify patterns with 95-100\% accuracy for $\alpha \leq 1$. Further, the capacity reduces to about 80\% as the load on the system increases to $\alpha=2$.

We can also see that performance of the network on the classification task is better than that on the memory task. As shown in Figures \ref{fig:MemCap} and \ref{fig:ClassCap}, the delay trained network can memorize about 50 random spike patterns with 90-100\% accuracy while it can classify about 140 patterns belonging to two classes with similar accuracy levels. The reason for this is that the memory task can be considered as a classification task where the entire background population is the second class. Hence, it is natural that the network will mis-classify more patterns.

\begin{figure}[!t]
\centerline{\includegraphics[width=1\textwidth]{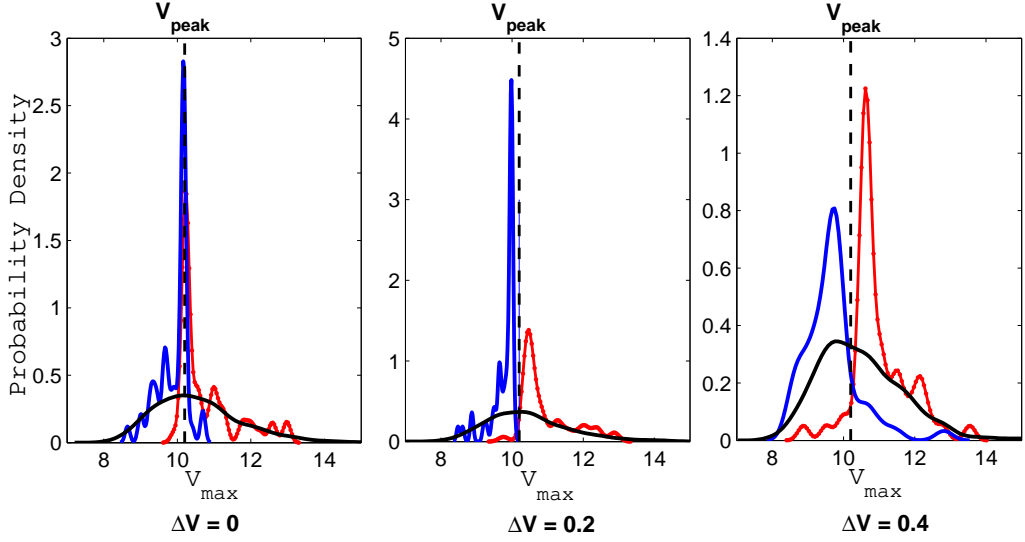}}
\caption{Probability density distribution of $V_{max}$ for $50$ patterns learnt to be in class 1 (red) and in class 2 (blue). The effect of learning using different $\Delta V$ values on the $V_{max}$ distributions for the two classes is shown. The $V_{max}$ distribution for new unseen patterns presented to the trained network is shown in black. Dashed vertical lines denote the location of $V_{peak}$.}
\label{fig:Class_Dist}
\end{figure}

\begin{figure}[!t]
\centerline{\includegraphics[width=0.6\textwidth]{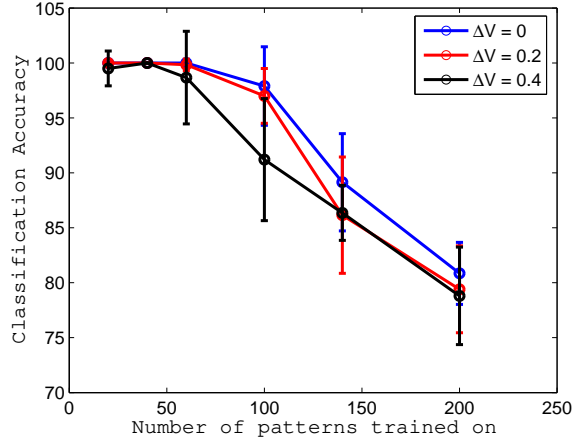}}
\caption{Classification accuracy of the model. The network was trained to classify random patterns belonging to class 1 and class 2. The fraction of correctly classified patterns belonging to both classes is a function of the total number of patterns the network is trained on and the threshold $V_{thr}$ used for each class.}
\label{fig:ClassCap}
\end{figure}

\begin{figure}[!t]
\centerline{\includegraphics[width=0.6\textwidth]{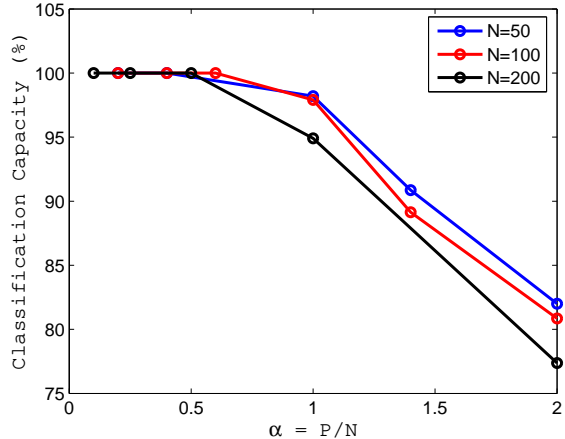}}
\caption{Capacity of the model as a function of the load on the system. The network was trained on $P$ patterns for different number of synapses $N=50, 100$ and $200$, $\Delta V=0$. The classification accuracy is plotted against $\alpha=P/N$. The capacity of the model reduces as $\alpha$ increases.}
\label{fig:Capacity_Load}
\end{figure}

\begin{figure}[!t]
\centerline{\includegraphics[width=0.6\textwidth]{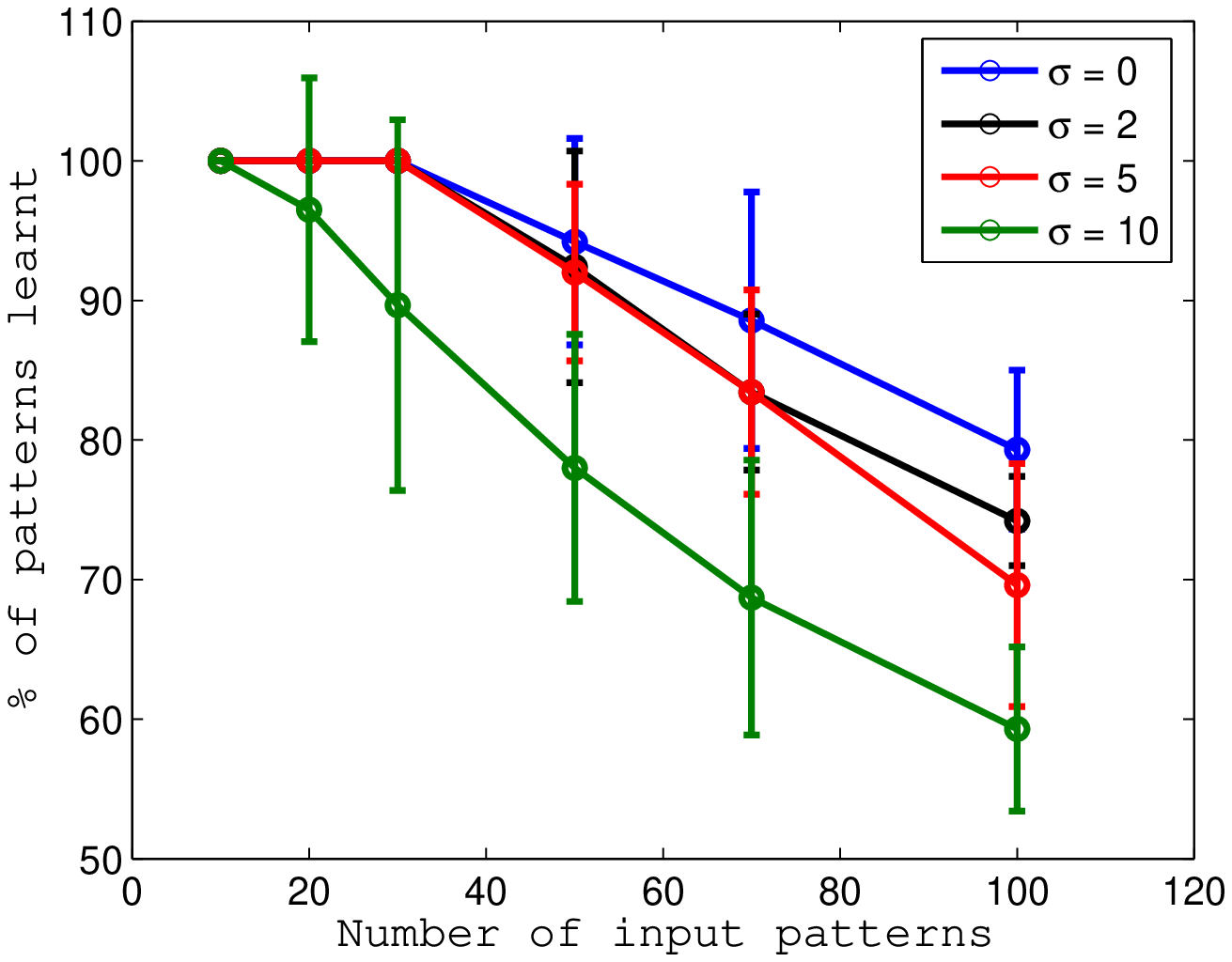}}
\caption{Memory capacity of the model in the presence of noise in $t_{max}$ estimate. A jitter of $\sigma = 2, 5$ and $10$ ms was added to the $t_{max}$ values. The pattern memorization capacity drops as higher noise level is introduced by increasing $\sigma$. $V_{thr}=10.7$.}
\label{fig:tmaxJitter}
\end{figure}

\item[(IV)] \textbf{Effect of non-idealities}\\
We also tested the noise tolerance of the delay learning scheme. These experiments tested the degree of resilience of our model to the overfitting problem of a pattern memorization task. First, we introduced noise in the $t_{max}$ estimate by adding Gaussian noise with zero mean and standard deviation $\sigma$ ms. We computed the pattern memorization capacity of the network using the delay learning scheme in which noisy estimates of $t_{max}$ were used to modify delays. As shown in Figure \ref{fig:tmaxJitter}, memory capacity drops in the presence of noise in $t_{max}$ as compared to the case when ideal estimate of $t_{max}$ is used ($\sigma=0$). As $\sigma$ is increased, the estimate of $t_{max}$ drifts further away from its ideal value, leading to a more difficult and time consuming learning process. Hence, for small number of patterns and moderate values of $\sigma$ ($\sim 5$ ms) the capacity is almost unchanged and only when $\sigma$ is close to the synaptic current fall time constant $\tau$, the capacity degrades drastically.

Next, we estimated the capacity of our delay learning model to recall patterns when presented with incomplete or noisy versions of the trained input patterns. Jittered versions of spike patterns were generated by adding Gaussian noise to each spike time. A jitter of 1.5 ms was added to all the spike patterns. The ability of the network to recognize noisy patterns was determined by calculating the memory capacity in the same way as before. The network was first trained on random spike patterns. In practice, it is useful to set the threshold during recall to be slightly less than $V_{opt}$ so that we can recognize jittered patterns at the cost of slightly increased FP errors. So we have set threshold to the minimum of ($V_{thr}-0.2$) and $V_{opt}$ in order to calculate how many jittered versions of the trained unperturbed spike patterns are recalled by the network. For small to moderate number of patterns ($P=10-50$) and $V_{thr}=10.7$, $V_{opt}$ estimated is usually higher than ($V_{thr}-0.2$) and therefore we use this lower threshold for recalling noisy patterns. This introduces small FP errors in the range of 0.5-2\%. For larger number of patterns and higher $V_{thr}$, it is found that $V_{opt}$ is much lower than $V_{thr}$ and therefore the choice of $V_{opt}$ to recall jittered patterns seems reasonable with no further increase in the FP errors.

\begin{figure}[!t]
\centerline{\includegraphics[width=0.6\textwidth]{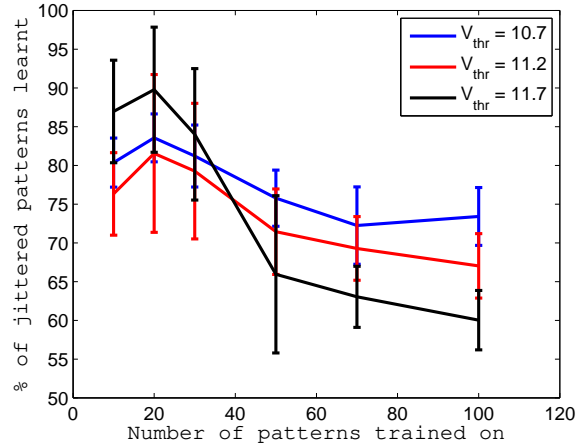}}
\caption{Memory capacity of the model in the presence of temporal noise. Gaussian noise with standard deviation 1.5 ms was added to all spike times. The ability of the model to recognize these noisy patterns reduces by 5-15\% from its memorization capacity for unperturbed spike patterns.}
\label{fig:JitMemCap}
\end{figure}

As shown in the Figure \ref{fig:JitMemCap}, the memory capacity drops by about 5-15\% in the presence of temporal noise. The reduction in memorization capacity for noisy patterns is higher for lower $V_{thr}$ ($=10.7$) while the capacity doesn't deteriorate by much for higher threshold ($V_{thr}=11.7$). For example, when trained on 100 spike patterns, the capacity reduces by about 10\% for $V_{thr}=10.7$ while it reduces by only about 4\% for $V_{thr}=11.7$. The reason for this phenomenon can be explained by the fact that the separation between the $V_{max}$ distributions for learnt and background spike patterns is small for low $V_{thr}$ as shown in Figure \ref{fig:ProbDist}. A jittered version of the trained spike pattern has a higher chance of falling in the $V_{max}$ distribution for the background patterns at lower $V_{thr}$ than at higher $V_{thr}$. Therefore, higher number of noisy versions of learnt unperturbed patterns become unrecognizable at lower thresholds.

Finally, we evaluated the performance of the model when presented with incomplete versions of the trained input patterns. Figure \ref{fig:MissingSp} shows the percentage of patterns memorized as a function of the number of spikes missing in the spike pattern. The network was trained on 50 random spike patterns consisting of 100 spikes and $V_{thr}=11.7$ and then 1, 2, 3, 4 and 5\% of afferents were randomly selected such that no spikes arrive at these afferents. The threshold for recalling patterns is set to min[($V_{thr}-0.2$), $V_{opt}$] as in the last experiment. As we can see that the memory capacity reduces from the case when complete patterns were being recalled (0\% missing spikes), from a mean value of 74\% of complete patterns to about 70\% of incomplete patterns, when 1 spike is missing. This drop in the network capacity when recognizing incomplete patterns is related to the probability that the missing spikes arrive close to the $t_{max}$ corresponding to each pattern and also to the difference ($V_{max}-V_{thr}$) for that pattern. Hence, the reduction in memory capacity for 1-2 missing spikes is not very significant. As more number of spikes are removed, the patterns become unrecognizable and the capacity reduces further.
\end{itemize}

\begin{figure}[!t]
\centerline{\includegraphics[width=0.6\textwidth]{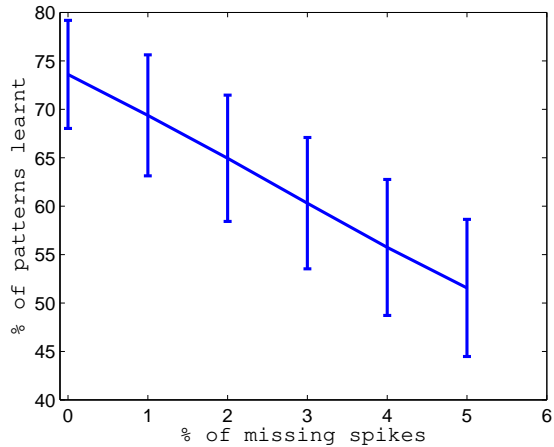}}
\caption{Memory capacity of the model for incomplete spike patterns. The capacity of the network to recall 50 patterns consisting of spikes missing at some of the $N=100$ afferents is shown. $V_{thr}=11.7$.}
\label{fig:MissingSp}
\end{figure}

\section{Hardware Architecture and Results}
\subsection{Architecture}
\begin{figure}[!t]
\centerline{\includegraphics[width=0.8\textwidth]{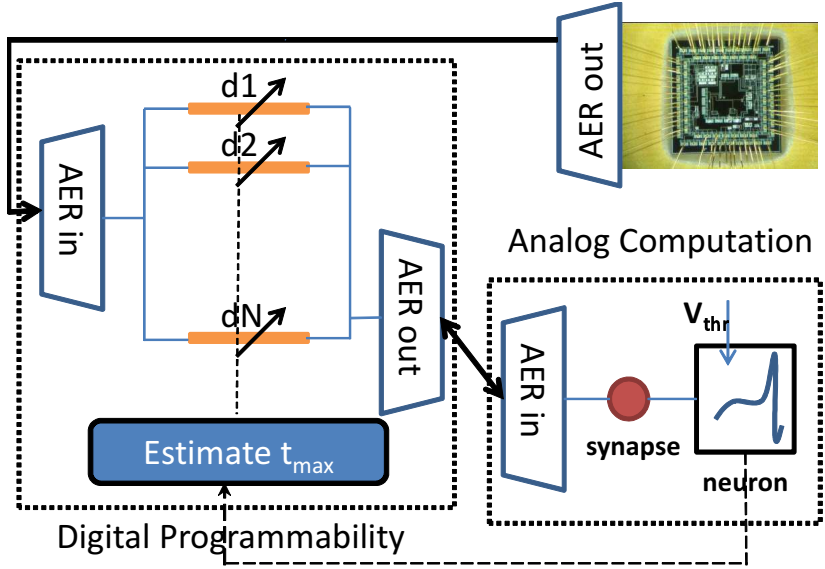}}
\caption{System architecture of the proposed hardware implementation of the DELTRON algorithm uses digital for learning and analog for computation.}
\label{fig:sysArch}
\end{figure}

We propose an efficient mixed-signal VLSI implementation of the DELTRON algorithm. Figure \ref{fig:sysArch} depicts the system level view of our proposed system where a spiking sensor communicates patterns to be memorized by our network. The network uses digital tunable delay lines as the memory storage element and communicates output spikes to an analog chip that houses a synapse and a spiking neuron. The inter-module communication can be handled by the AER protocol \cite{Mahowald1992}. During the learning process, another digital block is needed to estimate $t_{max}$ from the output spike train of the neuron (the connection is indicated by a dashed line in the figure). It should be noted that since we want to estimate $V_{max}$ from the output spike train, we have to ensure that the neuron generates spikes for any value of $V_{max}$. Hence, during the learning phase, the value of the threshold of the integrate and fire neuron is kept at zero. An advantage of this algorithm is that synaptic mismatch and DACs to implement weights are avoided. All the digital blocks, described in the following paragraphs, are also relatively simple.

\begin{itemize}
\item[(1)] Time-keeper: A global counter is kept which counts up till the maximum value of pattern time which is 400 ms in this case. The corresponding digital word will be referred to as $T_s$ and indicates the system time.
\item[(2)] Delay line: Every axon has two registers, R1 and R2. While R1 stores the current delay value $d$, R2 stores the value of $(x+d)$ which is the time when it should generate a spike. Whenever there is an incoming spike, R2 will be updated to the sum of $d$ and $T_s$, which equals $x$ at that time instant. At every clock the value in R2 is compared with the current time $T_s$, and if found same, the axon fires an output event request to the $AER_{out}$ module.
\item[(3)] Estimate $t_{max}$: This block comprises two counters, C1 and C2, and two registers to store the current estimate of $V_{max}$ and $t_{max}$ respectively. We estimate $V_{max}$ by measuring the inter-spike intervals (ISI) \--- a large $V_{max}$ corresponds to a small ISI and vice versa.  At the beginning of every cycle, C1, C2 and the $t_{max}$ register are reset to zero while the ISI register is reset to the maximum time. Every odd spike resets and starts C1 while stopping C2, while every even spike does the opposite. The counter that is stopped holds the current ISI which is compared with the current minimum ISI stored in the ISI register. If the current ISI is smaller, it is stored in the ISI register while $t_{max}$ is updated as $T_s$ - ISI/2.
\item[(4)] Learning module: After every input pattern presentation is over, the learning module will update the delays before the presentation of the next pattern. First, every R2 register is updated to the difference of $t_{max}$ and its current value (computing $t_{max}$ - $t_i$ in step (5) of the algorithm in Section \ref{sec:algo}). Only those axons with a positive value in R2 after this step get updated. The update is done serially by using the value in R2 of a chosen axon to index a look-up table that stores the values of $\eta K'(t)$. The value in R1 is modified by adding the value from the LUT to its current value.
\end{itemize}

\subsection{Hardware Simulations Results}
\begin{figure}[!t]
\centerline{\includegraphics[width=0.6\textwidth]{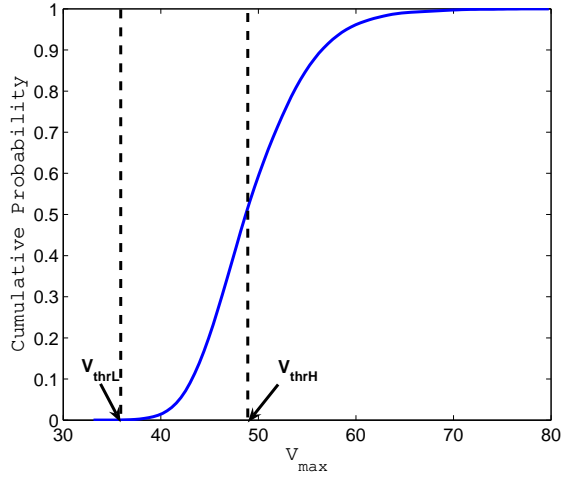}}
\caption{Cumulative distribution function of the $V_{max}$ of the I\&F neuron. The thresholds $V_{thrL}$ and $V_{thrH}$ are set as shown.}
\label{fig:V_InF_dist}
\end{figure}

It is not obvious that the delay learning algorithm will work well with the $t_{max}$ being estimated from the output spike train of the integrate and fire neuron. Since, it is computationally infeasible to run a co-simulation of the DELTRON learning algorithm with SPICE, hence we present the results of the learning algorithm for the hardware system from behavioral simulations performed in MATLAB. The learning algorithm used in Section \ref{sec:algo} relied on estimating $t_{max}$ from the membrane voltage $V(t)$ generated by summed EPSPs. Since, we cannot access $V(t)$, we have proposed estimating $t_{max}$ from the spike train output of the I\&F neuron. Here, we test the capacity of delay learning algorithm based on the new $t_{max}$ estimates to memorize random spike patterns. For this experiment we considered that a spike pattern $\bf x$ = ($x_1$, $x_2$, ..., $x_N$) arriving at $N$ synapses, generates excitatory postsynaptic currents (EPSCs) which get delayed by $\bf d$ = ($d_1$, $d_2$, ..., $d_N$). The total current due to all incoming spikes in a pattern is given by:
\begin{align}
I(t)&=\sum_{t_{i}}K(t-t_{i})\\
&=I_0\sum_{t_{i}}(exp[-(t-t_i)/\tau] - exp[-(t-t_i)/\tau_s])
\end{align}
where $K$ is the EPSC kernel as shown in Figure \ref{fig:Kernel} (top), $t_i=x_i+d_i$, normalization factor $I_0=2.12$ nA. All the other parameters $\tau$ and $\tau_s$ are set to the same values as before. The total input current received by the neuron generates membrane voltage ($V$) calculated using the leaky integrate and fire neuron model with membrane time constant $\tau_n=5$ ms and threshold voltage $V_{thr}$. The output neuron fires a spike if $V>V_{thr}$ at a time $t_{spk}$. After a spike, the membrane voltage $V$ is reset to zero. An input pattern is said to be memorized if the network generates output spikes in response to that pattern. The learning scheme used is briefly discussed next. A set of $P$ random spike patterns are generated and delays are initialized. The spike times $x_i$ are randomly drawn from a uniform distribution $[1$ $400]$ ms and delay $d_i$ for the i-th afferent is initialized between 0 and 50 ms, where i = 1, 2, ..., N and $N=100$. During training, $V_{thr}$ is set to a small value $V_{thrL}=36$ mV, such that all the untrained input patterns drive the output neuron to fire. Figure \ref{fig:V_InF_dist} shows the cumulative distribution plot of the maximum value of the membrane voltage of the I\&F neuron, $V_{max}$. It can be seen that probability $P(V_{max}\leq36) \approx 0$. Therefore, by setting $V_{thrL}=36$ mV we can ensure that $V_{max}>V_{thrL}$ and the output I\&F neuron fires for all the untrained input patterns. The higher threshold $V_{thrH}$ is the one used in actual learning; it is chosen in the same way as in the earlier sections. The delay-modification involves the following steps:
\begin{itemize}
\item[(1)] A spike pattern is presented to the network. The threshold voltage of the neuron is set to $V_{thrH}$.
\item[(2)] If the output neuron fires, the spike pattern is learnt. Delays are not modified and the next input pattern is presented.
\item[(3)] If the output neuron fails to fire, the spike pattern is not learnt. Threshold voltage is set to the lower threshold $V_{thrL}$ such that the output neuron generates spikes in response to this input pattern.
\item[(4)] $t_{max}$ is estimated from the output spike train by calculating the minimum $ISI=t_{spk2}-t_{spk1}$ and setting
\begin{equation}
t_{max} = \frac{(t_{spk1} + t_{spk2})}{2}
\end{equation}
\item[(5)] Delay changes are calculated as before according to $\Delta d_i = K'(t_{max}-t_i)$ where $K'$ indicates the derivative of EPSC kernel $K$.
\item[(6)] Threshold voltage is set to $V_{thrH}$. The new delay values are used to calculate the number of patterns memorized by the network. Delays are modified only if the number of learnt patterns increases.
\end{itemize}

\begin{figure}[!t]
\centerline{\includegraphics[width=0.8\textwidth]{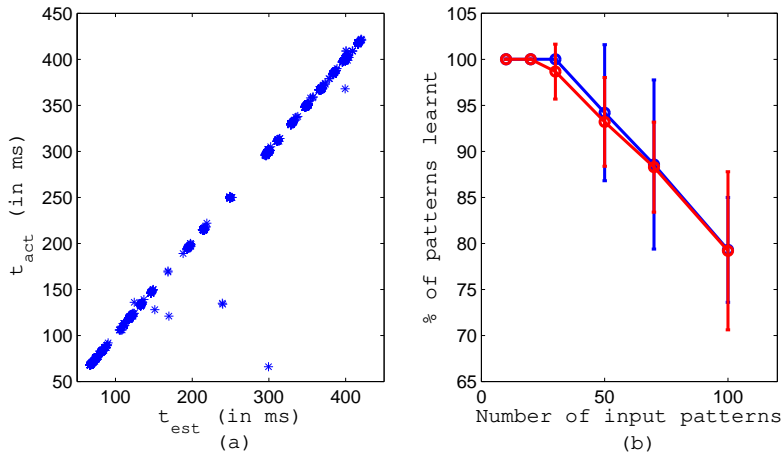}}
\caption{Results of the learning algorithm for hardware implementation. (a) Actual $t_{max}$ in each learning iteration from $V(t)$ (y-axis) and estimated $t_{max}$ from spike train output (x-axis). The two estimates are very similar. (b) Pattern memorization capacity using the two learning algorithms. Results of Section \ref{sec:simulation}(II) (blue) and learning algorithm for hardware system (red) show similar performance.}
\label{fig:IntFire}
\end{figure}

The stopping conditions for the learning process are same as discussed in steps (8) and (9) of the learning algorithm in Section \ref{sec:algo}. Figure \ref{fig:IntFire}(a) shows the comparison between the $t_{max}$ estimates obtained in each learning iteration of the two learning methods. We can see that the $t_{max}$ computed from $V(t)$ as in Section \ref{sec:algo} ($t_{act}$) and from spike train output ($t_{est}$) are very similar. The cases for which these two values are significantly different, as shown by the data points away from the `y=x' line, it was found that $V(t)$ was very similar at these two $t_{max}$ locations and therefore translated into similar bursts of output spikes around these times. Therefore, both the $t_{max}$ estimates were valid for pattern memorization task. Further, we have shown that the $t_{max}$ estimated from spike train output results in pattern memorization capacity similar to that obtained in Section \ref{sec:simulation}(II). As seen in the Figure \ref{fig:IntFire}(b), the memory capacity plots corresponding to the two $t_{max}$ computation methods are very similar. These results support our proposed hardware implementation of the DELTRON algorithm.

\begin{figure}[!t]
\centerline{\includegraphics[width=0.8\textwidth]{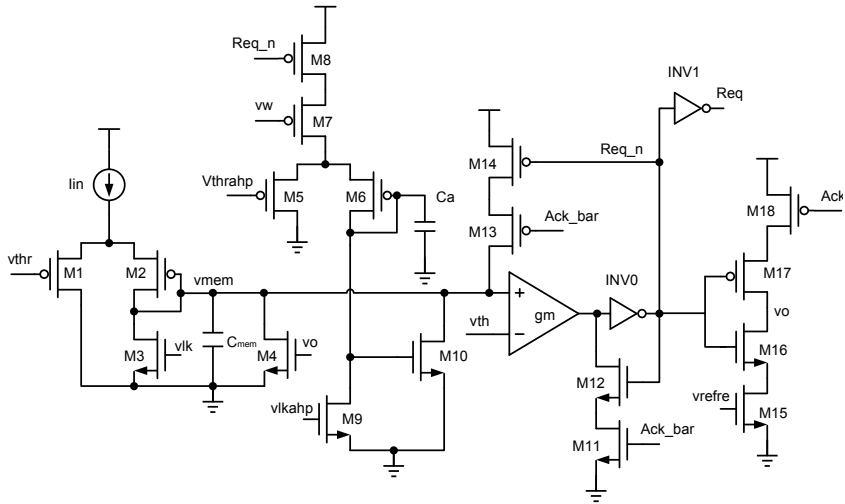}}
\caption{The neuron circuit is a modified version of the DPI neuron with an added transconductor based comparator for explicitly setting a voltage threshold.}
\label{fig:dpi_neuron}
\end{figure}

As mentioned earlier, a non-trivial step in the algorithm is the estimation of $V_{max}$ from the spike train output of an integrate and fire neuron. The rest of the circuits are based on well known digital circuits and are not discussed further. We have performed SPICE simulations of this part of the algorithm using transistor models from the AMS 0.35 um CMOS process. We used an approximation to a conductance based neuron model \cite{giacomo_neuron_conductance} for this simulation, though other neuron circuits can also be used\cite{nullcline_neu_my,Indiveri2011}. Figure \ref{fig:dpi_neuron} shows the circuit schematic of this neuron. We modified the original structure by including an operational transconductance amplifier (OTA) as an explicit comparator setting a well-defined value of $V_{thr}$. The part of the circuit for spike frequency adaptation was turned off by setting the bias voltages `vlkahp' and `Vthrahp' to a high and low value respectively. An extra positive feedback loop comprising `INV0' and `M12' is added to allow for rapid discharge of the output voltage of the OTA during reset. The synapse model used in the simulation is based on the differential pair integrator structure described in \cite{Bartolozzi2007} and is not described here.

\begin{figure}[!t]
\centerline{\includegraphics[width=0.8\textwidth]{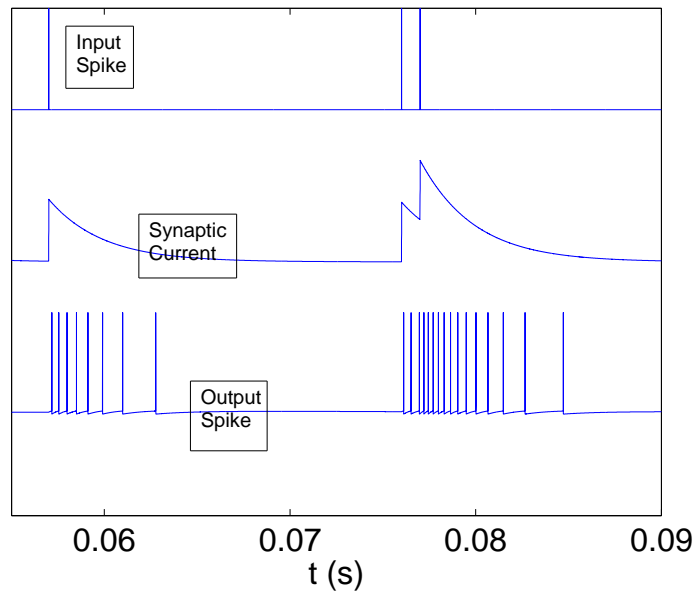}}
\caption{Current input waveform to the  integrate and fire neuron and ISI of the output spike train in SPICE simulations show that higher the current, smaller is the ISI and vice versa. Absolute values are not shown for ease of viewing.}
\label{fig:inputCurrent}
\end{figure}

Figure \ref{fig:inputCurrent} plots an example of the input current waveform for an input spike train consisting of 10 spikes at random times. The output spikes from the neuron  can be seen to provide a good estimate of the input current \--- larger the input current, smaller is the ISI or equivalently higher is the density of spikes. This validates our earlier assumption that ISI can be used to estimate $V_{max}$ (in this case it is equivalent to $I_{in,max}$).

\begin{figure}[!t]
\centerline{\includegraphics[width=0.6\textwidth]{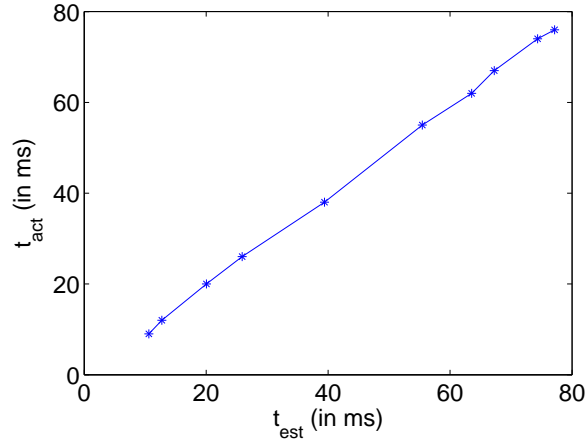}}
\caption{A plot of the estimated and actual $t_{max}$ values exhibit a high degree of correlation in SPICE simulations.}
\label{fig:tactPlot}
\end{figure}

Furthermore, Figure \ref{fig:tactPlot} plots the relationship between $t_{est}$, (estimated $t_{max}$) and $t_{act}$ (actual $t_{max}$) for 10 different random spike trains where the input consists of 10 spikes at random times between 0 and 100 ms. The high degree of correlation of these two value again validate the choice of ISI as a metric for $I_{in,max}$. As mentioned earlier, the threshold was kept at a very small value for this part. It should be noted that in this case $I_{leak}$ is the effective threshold since we are sensing the maximum of the input current.

\section{Conclusion}
We presented a new learning algorithm for spiking neural networks that modifies axonal delays (instead of synaptic weights) to memorize or classify spatio-temporal spike patterns. Although we have demonstrated the use of our model in a binary classification problem, its application can be extended to multiclass problem by having multiple DELTRONs. This is similar to the one-versus-all or one-versus-one approach used by Support Vector Machines (SVMs) for multiclass classification. The training algorithm uses a philosophy of updating as few axonal delays as possible to memorize a single pattern thus preserving memory capacity. The capacity of the network with $100$ axons for classification with accuracy greater than $90\%$ is around 100 patterns. It can classify patterns with about 80-100\% accuracy for $\alpha \leq 2$. For the tougher memorization task where the entire background population can be considered as a second class, it can memorize about 50 patterns with $90\%$ accuracy of recall. We also presented a mixed-signal VLSI implementation of the algorithm which requires only one tunable parameter, $V_{thr}$, depending on the statistics of the input. In contrast to this, most of the neuromorphic ICs have multiple tunable parameters. This delay learning implementation is simpler than the weight modification algorithms. However, the capacity of this algorithm is lesser than weight modification ones because modifying delays can increase $V_{max}$ by a limited amount only. Hence, a possible future avenue of research is to explore a combination of the two methods.

\section{Acknowledgements}
The authors would like to acknowledge the contribution and support of UNSW International Contestable Funding Scheme ``Building Asia-Pacific Collaborations in Neuromorphic Research" and also thank Shoushun Chen, Zohair Ahmed, Roshan Gopalakrishnan and Subhrajit Roy for useful discussions and help with SPICE simulations.
\bibliographystyle{elsarticle-num}

%\bibliography{reflist1}

\end{document}